\documentclass[
]{ceurart}

\usepackage{listings}
\begin{document}

\copyrightyear{2023}
\copyrightclause{Copyright for this paper by its authors.
  Use permitted under Creative Commons License Attribution 4.0
  International (CC BY 4.0).}

\conference{In A. Martin, K. Hinkelmann, H.-G. Fill, A. Gerber, D. Lenat, R. Stolle, F. van Harmelen (Eds.), 
Proceedings of the AAAI 2023 Spring Symposium on Challenges Requiring the Combination of Machine Learning and Knowledge Engineering (AAAI-MAKE 2023), Hyatt Regency, San Francisco Airport, California, USA, March 27-29, 2023.}

\title{Robot Behavior-Tree-Based Task Generation with Large Language Models}

\tnotemark[1]
\tnotetext[1]{This work was supported in part by the National
Science Foundation under Grant IIS-1813935.
Any opinion, findings,
and conclusions or recommendations expressed in this material are
those of the authors and do not necessarily reflect the views of
the National Science Foundation.}

\author[1]{Yue Cao}[%
email=yuecao@purdue.edu,
]
\cormark[1]
\address[1]{Elmore Family School of Electrical
and Computer Engineering, Purdue University, West Lafayette, IN 47907, U.S.A.}

\author[1]{C.S. George Lee}[%
email=csglee@purdue.edu,
]
%

\cortext[1]{Corresponding author.}

\begin{abstract}
 Nowadays, the behavior tree is gaining popularity as a representation for robot tasks due to its modularity and reusability. Designing behavior-tree tasks manually is time-consuming for robot end-users, thus there is a need for investigating automatic behavior-tree-based task generation. 
Prior behavior-tree-based task generation approaches focus on fixed primitive tasks and lack generalizability to new task domains. 
To cope with this issue, we propose a novel behavior-tree-based task generation approach that utilizes state-of-the-art large language models. 
We propose a Phase-Step prompt design that enables a hierarchical-structured robot task generation and further integrate it with behavior-tree-embedding-based search to set up the appropriate prompt. 
In this way, we enable an automatic and cross-domain behavior-tree task generation. 
Our behavior-tree-based task generation approach does not require a set of pre-defined primitive tasks. End-users only need to describe an abstract desired task and our proposed approach can swiftly generate the corresponding behavior tree. A full-process case study is provided to demonstrate our proposed approach. An ablation study is conducted to evaluate the effectiveness of our Phase-Step prompts. Assessment on Phase-Step prompts and the limitation of large language models are presented and discussed.
\end{abstract}

\begin{keywords}
 Robot task \sep
 Large language model \sep
 GPT-3 \sep
 Behavior tree \sep
 Task planning
\end{keywords}

\maketitle

\section{Introduction}

The behavior tree is a graphical representation of robot tasks. It originated from the video game industry and then is introduced to the robotics community~\cite{ghzouli2020behavior}. The behavior tree coordinates robot primitive tasks using a variety of control-flow nodes. Compared with traditional task representations such as state machines, subsumption architectures and decision trees, the behavior-tree formalism exhibits several advantages in terms of modularity, flexibility and readability. Some pioneering works have been conducted to apply behavior trees to robot tasks such as assembly~\cite{styrud2022combining} and navigation~\cite{macenski2020marathon}.

Designing behavior-tree-based robot tasks remains a cumbersome work for end-users. In order to improve the efficiency of the design process, several behavior-tree generation approaches have been proposed recently. Provided with a set of linear temporal logic specifications, a behavior tree can be constructed using designed conversion rules~\cite{colledanchise2017synthesis}. Alternatively, based on the primitive-task sequence produced from the PDDL planner, a behavior tree can be created to achieve the optimal-task execution time~\cite{rovida2017extended}. In addition, behavior trees can also be generated with the integration of heuristic search~\cite{scheide2021behavior} or reinforcement learning~\cite{kartavsev2021improving}. However, all these approaches rely on a pre-defined library of primitive tasks, and none of them is capable of creating behavior trees that are out of given domains. For example, if primitive tasks, such as ``align wheel'', ``weld door'', and ``place enginee'', 
do not exist in the pre-defined library, it is impossible to generate a behavior-tree task for car assembly.

Recent advances in large language models (LLMs) offer a potential direction to improve in generalizability of robot-task generation. The large language models are neural-language models with a large number of parameters and trained over huge amount of data. Such large language models show powerful generalizability in many natural language processing (NLP) tasks. Starting from the GPT-3 model~\cite{brown2020language} in 2020, large language models are becoming an emerging research area in natural language processing and they also draw attention from robotics researchers. 

\begin{figure}[h]
    \centering
    \includegraphics[width=\linewidth]{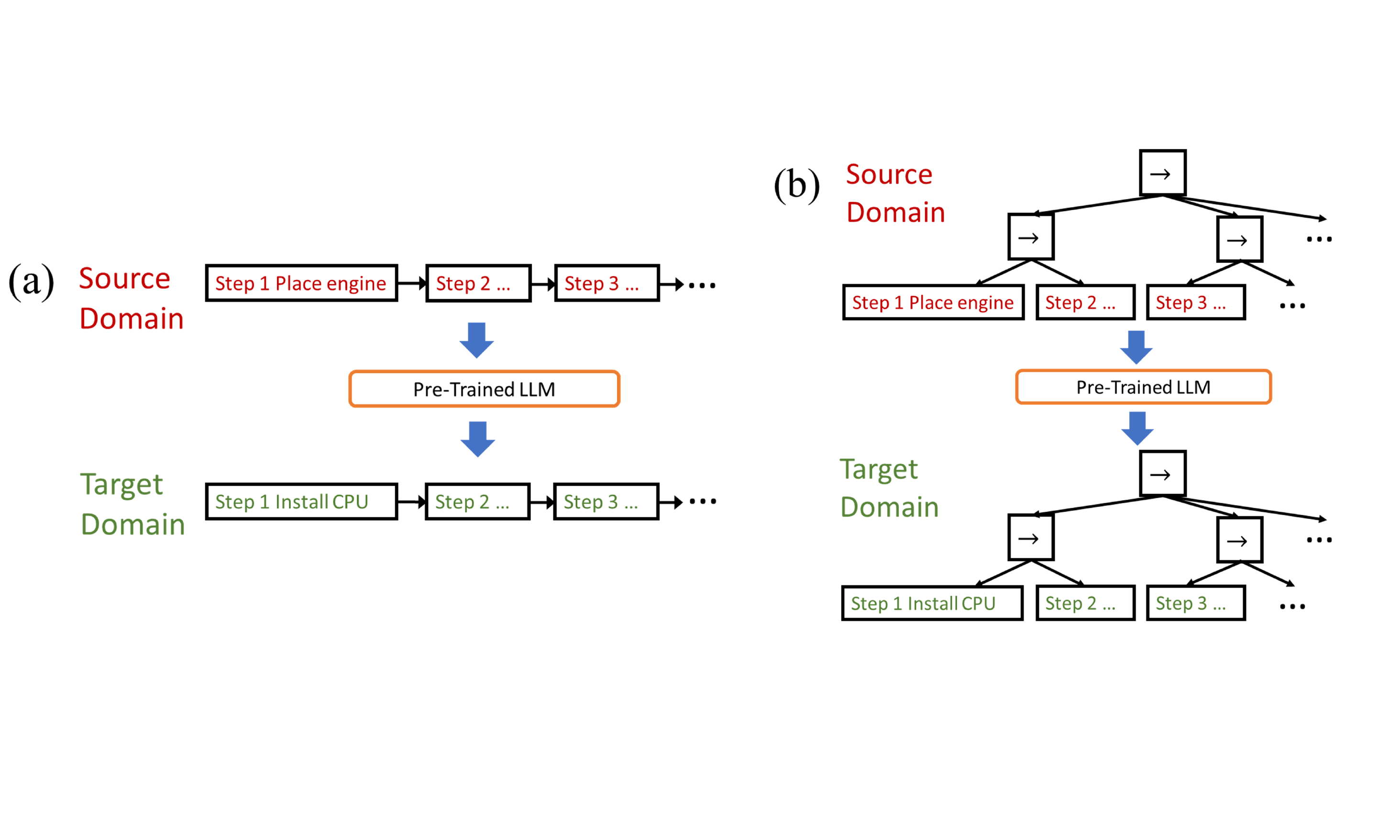}
    \caption{Cross-domain task generation using a pre-trained large language model (Source domain: Car assembly; Target domain: Computer assembly).  (a) Prior work generated sequential tasks; (b) We aim to generate modular tasks in behavior-tree form.}
    \label{fig:my_label1}
\end{figure}

Some recent works considered the autoregressive pre-trained LLM 
as a robot-task planner and used it to generate task sequences in a zero-shot or few-shot fashion. With the pre-trained LLMs, the robot-task-generation problem can be formulated in the prompt-based learning paradigm~\cite{liu2021pre} instead of the classical supervised-learning paradigm. In~\cite{pmlr-v162-huang22a}, an abstract task descriptor and its associated sequence of primitive tasks are used to set up the prompt. The SayCan framework~\cite{ahn2022can} combines a human high-level instruction and its corresponding robot primitive tasks as the prompt. The ProgPrompt~\cite{singh2022progprompt} represents robot tasks as Pythonic programs and then takes the Pythonic code as the prompt. Unfortunately, the existing works are limited to the sequential-task generation of robots. By contrast, modular-task representations, such as behavior trees, have better reusability and readability than the sequential-task representation. 
Hence, we aim for generating behavior-tree robot tasks using large language models, as shown in Figure~\ref{fig:my_label1}. 

In this paper, we propose a novel robot behavior-tree-task-generation approach using pre-trained large language models. We first design a Phase-Step prompt that enables a 3-layer behavior-tree generation. Next we introduce a method that generates a complete behavior tree from the initial 3-layer behavior tree. Afterwards, we utilize a behavior-tree task knowledge base storing a collection of robot tasks. We automatically query the knowledge base and select the behavior tree that is most similar to the desired task description as the prompt. In such process, a behavior tree can be generated for the desired task in a new domain.

Major contributions of this paper are:
\begin{enumerate}
\item[(1)] To our best knowledge, this is the first attempt to apply the state-of-the-art large language models to behavior trees. The powerful generalizability of LLMs enables robot-task generation in new domains rather than conforming to fixed primitive tasks. 
\item[(2)] Our proposed Phase-Step prompt extends the LLM-based task generation from sequential tasks to hierarchical tasks. By incorporating the modularity and reusability of behavior trees, it further eases the difficulty and burden of end-user design.
\end{enumerate}

\section{Background}
\subsection{Behavior-Tree Fundamentals}
\begin{figure}[h]
    \centering
    \includegraphics[width=0.8\linewidth]{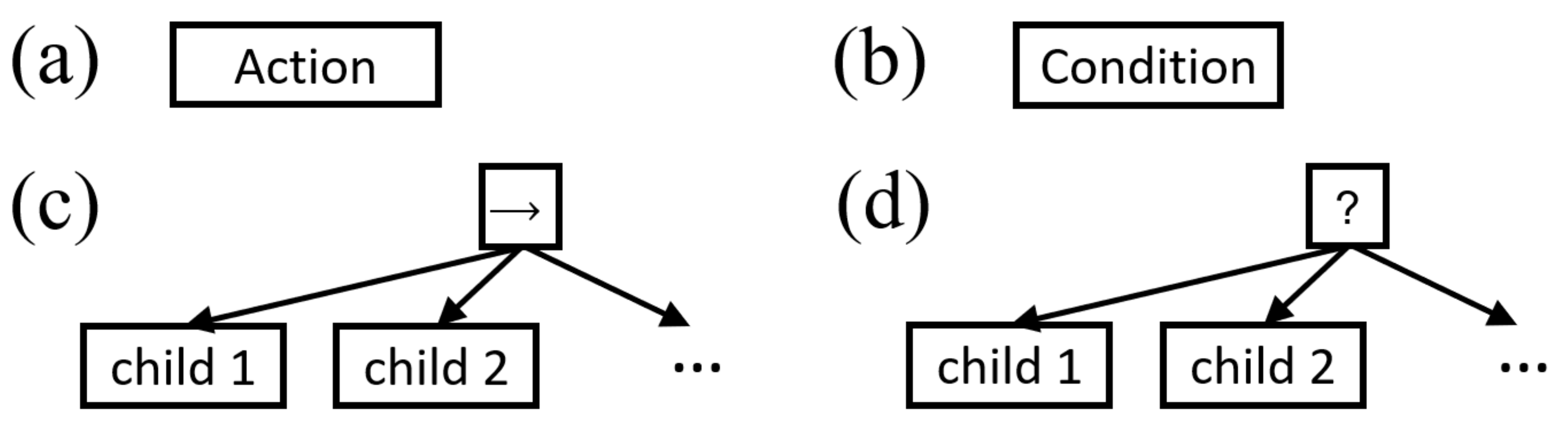}
    \caption{Graphical representation of four types behavior tree nodes. (a) {\em Action} node. (b) {\em Condition} node. (c) {\em Sequence} node. (d) {\em Fallback} node.}   
    \label{fig:background}
\end{figure}

The behavior tree is a hierarchical representation of robot tasks~\cite{ghzouli2020behavior}. Alternatively to the state machine, the behavior tree can be used to coordinate robot tasks at the abstract level. Formally, a behavior tree is defined as a directed rooted tree. The leaf nodes of the behavior tree determine the execution, and the branch nodes specify the control-flow logic~\cite{iovino2020survey}. The leaf nodes are either {\em Action} nodes or {\em Condition} nodes. The {\em Action} node specifies a certain robot primitive task and returns a {\em Success} when the primitive task is completed. The {\em Condition} node is used to evaluate the assigned Boolean condition, such as whether the robot-sensor reading satisfies the requirement. Two most commonly used branch nodes are the {\em Sequence} node 
and the {\em Fallback} node. The {\em Sequence} node runs all its child nodes sequentially until receiving the first {\em Failure} signal, 
and the {\em Fallback} node executes all its child nodes sequentially until receiving the first {\em Success} signal. With these four types of nodes, a behavior tree can achieve the same task execution as a state machine. 
Other branch node types, including the {\em Parallel} node and the {\em Decorator} node for extensible functionality are not discussed in this work. The graphical illustration of these four types of behavior-tree nodes is shown in Figure~\ref{fig:background}. 

The behavior tree is executed by an activation
signal called {\em Tick}. The {\em Tick} generates from the root node at a certain frequency and traverses the behavior tree following the control-flow nodes. Once a node is ticked, it returns its status to its parent node. Compared to traditional task representations such as state machines, the behavior-tree formalism offers several benefits, including modularity, flexibility and readability. Small behavior trees can be merged into a large behavior tree using control-flow nodes. Subtrees can be modified individually without taking external links into account. In contrast with dense nest representation of state machines, the succinct-tree representation is more readable towards end-users.

\subsection{Large Language Models and Prompt-Based Learning}
In the natural language processing, the text-generation problem can be described as a language model using conditional probability distribution,
$P(y|x; \theta)$, where $x$ is the input text, $y$ is the text to be generated, 
and $\theta$ are model parameters. 
In the era of deep learning, various deep neural networks have been utilized  as the language models. The neural-language models that contain a large number of parameters and are pre-trained over huge-scale corpus are referred as ~\textit{Large Language Models}. Nowadays, the large language models are empowered with strong generalizability over a variety of downstream natural-language tasks. 

Some state-of-the-art large language models also show strong capability in few-shot learning~\cite{brown2020language}. One example is the OpenAI GPT-3 model, which has 175 billion parameters. This few-shot learning capability shifts the traditional ``pre-train-then-fine-tune'' paradigm to the new ``prompt-based-learning'' paradigm~\cite{min2021recent}. The \textit{prompt} is the text inserted into the input, which modifies the original input into a template string with some unfilled slots and then guides the pre-trained model to perform a certain NLP task. Specifically, the original input text $x$ is firstly converted to a new form $x'$ with prompt function, $x'=f_{prompt}(x)$. The $x'$ is a new textual string consisting of 3 parts, the input slots [\enspace]$_X$, the output slots [\enspace]$_Y$, and additional prompt text. The input slots [\enspace]$_X$ accept $x$, and the output slots [\enspace]$_Y$ are open to be filled. The pre-trained large language model takes string $x'$ as its input and generates texts for the output slots. Then we map texts fill in [\enspace]$_Y$ to the final output $y$.   

In the prompt-based-learning paradigm, the prompt design plays a key role. It has been investigated that the prompt in GPT-3 mainly guides the language model to access the existing knowledge and regulates the output structure~\cite{reynolds2021prompt}.  For existing works using large-language-models for robot tasks, we summarize their prompt design in Table~\ref{tab:SumPrompt}. 

\begin{table}[htp]
\centering
\caption{Recent works of prompt design for robot-task generation. Task A and Task B refer to the source- and target-task description, respectively. [\enspace]$_X$ is the input slot and [\enspace]$_Y$ is the output slot.}
    
    \begin{tabular}{|c|p{100mm}|}
     \hline
        \textbf{Work}&\textbf{Prompt Design}   \\
          \hline LLM Planner~\cite{pmlr-v162-huang22a}
          & Task: Task A; Step 1: [Sub-task 1]$_X$; Step 2: [Sub-task 2]$_X$; ...
          \newline
          Task: Task B; [Step 1: Sub-task ?; Step 2: Sub-task ?; ...]$_Y$\\
         \hline 
         SayCan~\cite{ahn2022can}
         &  How would do Task A? 1. [Sub-task 1]$_X$, 2. [Sub-task 2]$_X$, ...
         \newline
         How would do Task B? I would: [1. Sub-task ?, 2. Sub-task ?, ...]$_Y$
         \\
         \hline
        ProgPrompt~\cite{singh2022progprompt} & def Task A(\thinspace): \# 1: [Sub-task 1]$_X$; \# 2: [Sub-task 2]$_X$; ...\newline def Task B(\thinspace): [\# 1: Sub-task ?; \# 2: Sub-task ?; ...]$_Y$ \\
          \hline
    \end{tabular}
    \label{tab:SumPrompt}
\end{table}

\section{Proposed Approach}
Our proposed approach for behavior-tree task generation using pre-trained large language models consists of 3 key components -- Phase-Step prompt, behavior-tree construction, and automatic source-task selection. The Phase-Step Prompt enables the generation of a 3-layer behavior-tree fragment. 
Taking the fragment generated by the Phase-Step prompt, the construction process grounds the natural language to robot actions and produces a complete behavior tree. The automatic source-task selection unitizes the task retrieval from a knowledge base and further reduces the burden of manually designing prompts. 

\subsection{Phase-Step Prompt}
\begin{figure}[h]
    \centering
    \includegraphics[width=0.7\linewidth]{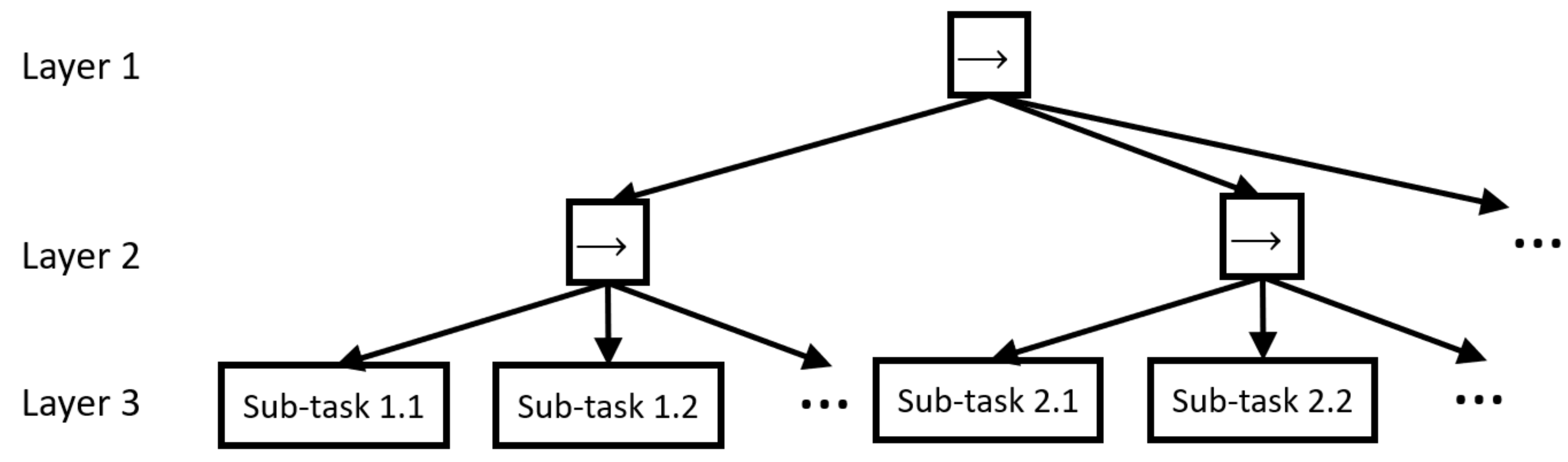}
    \caption{A 3-layer behavior tree consisting of {\em Sequence} nodes and {\em Action} nodes.}   
    \label{fig:tree3layer}
\end{figure}

Firstly, we introduce the \textit{Phase-Step Prompt}, 
which enables the generation of a 3-layer behavior tree. 
For any 3-layer behavior tree consisting of {\em Sequence} nodes and {\em Action} nodes as shown in Figure~\ref{fig:tree3layer}, we re-formulate it into a textual prompt in Table~\ref{tab:DesignPrompt}. 

The prompt starts with an auxiliary textual description ``Source Task'' and ``Procedures'', then it is followed by the detailed behavior-tree description for the source task. We refer the primitive task description in the $j$th {\em Action} node underneath the $i$th layer-2 {\em Sequence} node as ``Sub-task $i.j$''. 
For the $i$th layer-2 {\em Sequence} node, we represent it as ``Phase $i$'' in the prompt. For the {\em Action} node corresponding to `Sub-task $i.j$'', we place ``Step $j$''  after the ``Phase $i$'' in the prompt, then fill the textual description in the input slot as [Sub-task $i.j$]$_X$. After the source-task description, the prompt continues with an auxiliary text ``Target Task'' and our desired task description. The desired task description can be specified as a short phrase, such as ``Assembly a computer''. Optionally, we can also add the tool and the number of phases in the desired task description, such as ``Make coffee with coffee machine in 5 phases''. Once finishing the prompt specification, we apply the pre-trained LLM to generate the result in the output slot [\enspace]$_Y$. The generated result will also be in the Phase-Step form, which can be re-constructed into a new 3-layer behavior tree. 

\begin{table}[htp]
\centering
\caption{Phase-Step-prompt Design. [\enspace]$_X$ is the input slot and [\enspace]$_Y$ is the output slot.}
\begin{tabular}{l}
     \hline
     Source Task \\
Procedures: \\
Phase 1. \\
Step 1. {[}Sub-task 1.1{]}$_X$; Step 2. {[}Sub-task 1.2{]}$_X$; ...\\
Phase 2. \\ 
Step 1. {[}Sub-task 2.1{]}$_X$; Step 2. {[}Sub-task 2.2{]}$_X$; ...\\
...\\
\\
Target Task: Task Description\\
Procedures: \\
{[}Phase 1. \\
Step 1. Sub-task ?; Step 2. Sub-task ?; ...\\
Phase 2. \\ 
Step 1. Sub-task ?; Step 2. Sub-task ?; ...\\
...{]}$_Y$ \\
\hline
    \end{tabular}
    \label{tab:DesignPrompt}
\end{table}

\subsection{Behavior-Tree Construction}
We have presented how to use a Phase-Step prompt to generate a 3-layer behavior tree. Next, we discuss how to extend the generated 3-layer behavior-tree fragment into a complete behavior tree for robot tasks. 

In the generated 3-layer behavior-tree fragment, the sub-tasks may not be primitive tasks. In other words, the verb in the generated sub-tasks may not match the actual capability of robots, and thus the sub-tasks need further decomposition. In order to determine whether a sub-task is primitive, we create a verb list based on the robot capability, such as $L = \{\text{pick}, \text{drop}, \text{push}, \text{pull}, \text{rotate}, \text{move}, \text{place}\}$. This verb list can vary when using different types of robots. For each sub-task, we apply a language embedding model $Enc_1(\enspace)$ and perform an angular-cosine similarity computation:
\begin{equation}\label{eq:Sim_vector}
    \text{Sim}(v, L_i) =1 - 2 \arccos{(\frac{Enc_1(v)\cdot Enc_1(L_i)}{\lVert Enc_1(v) \rVert \lVert Enc_1(L_i) \rVert})}/\pi,
\end{equation}
where $v$ is the verb of a certain sub-task, $L_i$ is the verb in the verb list $L$ that corresponds to robot capability. If all the similarity scores associated with a certain sub-task are below a given threshold, we conduct a tree expansion over this sub-task. 
Specifically, we keep the source-task part in the prompt unchanged. We set the sub-task as the target-task description and enforce the number of phases to be 1. In such way, a sub-task can be decomposed into fine-grained sub-tasks. We continue performing such tree expansion until all the sub-tasks are primitive.

Another strategy to expand the generated behavior tree based on robot capability is directly modifying the prompt. If one sub-task is not primitive, we can set the target-task description in the prompt as ``sub-task... in 1 phase, only use the following verb: pick, drop, push, pull, rotate, move, place''. With this strategy, we can also make all sub-task primitive. This strategy reduces the computation but the expanded depth will be limited to 1. It can be used to freeze the behavior-tree expansion.

Moreover, generated sub-tasks sometimes contain extra specification attaching to the verb-object form, such as ``Move robot arm to desired location'' or ``Tighten clamps for stability''. In order to deal with the extra specification, we introduce the {\em Fallback} nodes and {\em Condition} nodes. For a certain sub-task with extra specification, we plug in a {\em Fallback} node above it in the behavior tree. Then we insert a {\em Condition} node as the left child of the new {\em Fallback} node. The extra specification fills the {\em Condition} node 
as an ``if ...'' statement, such as ``if reaches desired location'' or ``if stable''.

\subsection{Automatic Source-Task Selection}
So far, we have presented the process of generating a complete behavior tree using a pre-trained LLM. In the entire process, the source task used in the prompt plays an important role. The choice of the source task directly affects the quality of generated target task. Instead of manually selecting a source task, we propose an approach that enables automatic selection.

We consider a knowledge base storing a collection of robot behavior-tree tasks. From this knowledge base, we would like to choose a behavior-tree task that is semantically similar to the desired target task. For example, in order to prompt a desired ``Assembly computer'' task, we prefer using an ``Assembly car'' behavior-tree task instead of using a ``Make coffee'' behavior-tree task to prompt. To realize such similarity-based selection, we adopt the behavior-tree embedding proposed in~\cite{cao2022behavior}. The behavior-tree embedding produces a numerical vector for each behavior tree such that semantically similar behavior trees are encoded closely in the vector space. With the behavior-tree embedding, we can quickly retrieve the most similar behavior tree from the given knowledge base. In detail, for the desired target task, we put away its optional description including tool usage and phase numbers and only keep its core ``verb-object'' phrase. Then we apply a same sentence-embedding model $Enc_2(\enspace)$ used in the behavior-tree-embedding method and encode the target-task phrase $target$ into a vector $v_{target}$. Once we acquire the embedding of the target task, we compute the similarity between it and the embedding $u_i$ of every 3-layer behavior tree $BT_i$ in the knowledge base,
\begin{equation}\label{eq:Sim_vector2}
    \text{Sim}(target, BT_i) =  \frac{v_{target}\cdot u_i}{\lVert v_{target} \rVert \lVert u_i \rVert} .
\end{equation}
Afterwards, we perform similarity sorting and select the behavior tree with highest similarity as the source-task prompt.

\begin{figure}[h]
    \centering
    \includegraphics[width=1\linewidth]{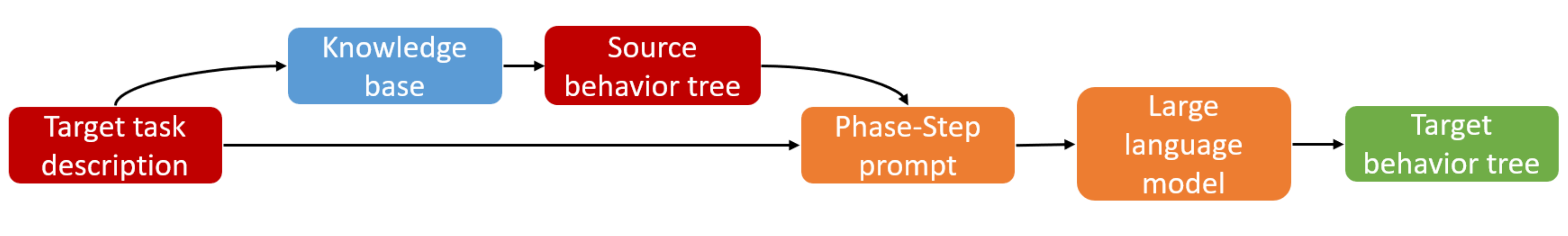}
    \caption{Proposed pipeline for robot behavior-tree task generation with large language models.}   
    \label{fig:pipe}
\end{figure}

In summary, the complete pipeline is shown in Figure~\ref{fig:pipe}. The generation process starts from a desired target-task description. The target-task description goes through a similarity sorting in the knowledge base and determines the most suitable 3-layer source behavior tree. Then both source behavior tree and target-task description are used to specify the Phase-Step prompt. The Phase-Step prompt is fed into the large language model and gradually generates a complete target behavior tree. The end-user only needs to set up the target-task description whereas the rest of the process is automatic.

\section{Case Studies and Evaluations}

\subsection{Experimental Setting}
We showcase our proposed approach using two state-of-the-art large language models from the OpenAI company -- GPT-3 text-davinci-003~\cite{davinci} and ChatGPT~\cite{chat}. Both large language models were built over the InstructGPT models~\cite{ouyang2022training} and initially released in November 2022. 

\textbf{GPT-3 text-davinci-003:} In order to obtain deterministic results, we configured the temperature parameter of 
GPT-3 text-davinci-003 model to be 0. We set the maximum length to be 500 tokens to allow the model to generate sufficiently long text. Another 3 parameters -- top P, frequency penalty, and presence penalty -- were set to 1, 0.2, and 0, respectively.

\textbf{ChatGPT:} The ChatGPT version used in our test is released on February 13, 2023. We used a ChatGPT Plus account and obtained results via its default online interface. Since the ChatGPT has not offered hyper-parameter configurations towards end-users and its generated results vary every time, we only recorded the results displayed for the first time. In addition, we started a new chat session for each target task rather than kept testing in the same session.

Regarding the two encoders $Enc_1(\enspace)$ and $Enc_2(\enspace)$ used in our approach, we adopt the universal sentence encoder~\cite{cer2018universal} for both. The universal sentence encoder is a general-purpose embedding model that can be applied to words, sentences, and paragraphs. We use the verb list $L = \{\text{pick}, \text{drop}, \text{push}, \text{pull}, \text{rotate}, \text{move}, \text{place}\}$ to ground natural language to robot actions. The similarity threshold used to determine the primitive actions was set to 0.5.

\subsection{Full-Process Demonstration}
Firstly, we presented a full-process demonstration of our approach. The approach performed similarity-based search in the knowledge base and selected the most suitable source target. For the knowledge base, we created a toy set consisting of four behavior-tree tasks. These four tasks ranged in four different categories including manufacturing assembly, logistics packaging, kitchen cooking, and household cleaning. Our desired target task is simply set as ``Desktop assembly''. The chosen source task was a wheel-assembly task~\cite{lange2010assembling} in the automotive manufacturing. Both GPT-3 text-davinci-003 and ChatGPT models took this wheel-assembly task to fill their prompts. The generated results for the initial 3-layer behavior-tree are shown in Table~\ref{tab:TestPrompt}.  The GPT-3 text-davinci-003 model generated a 4-phase task with 11 steps for completion, whereas the ChatGPT model generated a 3-phase task with 8 steps for completion. 

\begin{table}[htp]
\centering
\caption{The Phase-Step prompt is showcased using the GPT-3 text-davinci-003 and ChatGPT models. The same wheel-assembly task is used as the source task. The target task is generated inside the [\enspace]$_Y$ slot. }
\begin{tabular}{l|l}
     \hline
    GPT-3 text-davinci-003 & ChatGPT\\
     \hline
   Source Task &  Source Task\\
Procedures: & Procedures: \\
Phase 1. & Phase 1. \\
Step 1. Put car at a conveyor; &Step 1. Put car at a conveyor;\\
Step 2. Lift the car.&Step 2. Lift the car.\\
Phase 2. & Phase 2.\\ 
Step 1. Pick the wheel; & Step 1. Pick the wheel;\\
Step 2. Approach conveyor;& Step 2. Approach conveyor;\\
 Step 3. Align wheel with wheel hub. & Step 3. Align wheel with wheel hub. \\
Phase 3. & Phase 3. \\
Step 1. Insert screws; & Step 1. Insert screws; \\
Step 2. Fasten screws; &Step 2. Fasten screws; \\
Step 3. Leave the conveyor. & Step 3. Leave the conveyor.
\\
\\
Target Task: Desktop assembly & Target Task: Desktop assembly\\
Procedures: & Procedures: \\
{[}Phase 1. & {[}Phase 1. \\
Step 1. Place the desktop on a flat surface;& Step 1. Place desktop case on table;\\
Step 2. Connect the power cable to the power supply. & Step 2. Insert motherboard into the case.\\

Phase 2. & Phase 2. \\ 
Step 1. Attach the CPU to the motherboard; & Step 1. Install CPU; \\
Step 2. Connect the CPU fan to the motherboard; &  Step 2. Install RAM;\\
Step 3. Secure the CPU with screws. & Step 3. Install power supply.\\
Phase 3. &  Phase 3.\\
Step 1. Connect the RAM to the motherboard; &Step 1. Connect all cables and peripherals;\\
Step 2. Secure the RAM with screws;&  Step 2. Power on the device;  \\
Step 3. Connect the graphics card to the motherboard. & Step 3. Test for proper functionality.{]}$_Y$\\
Phase 4. & \\
Step 1. Connect the hard drive to the motherboard; & \\
Step 2. Secure the hard drive with screws; & \\
Step 3. Connect all cables to their respective ports &  \\
on the motherboard.{]}$_Y$ & \\
\hline
    \end{tabular}
    \label{tab:TestPrompt}
\end{table}

Next, a verb-list-based similarity computation was conducted to examine whether the generated action was primitive. According to the computation, all verbs appearing in the GPT-3 text-davinci-003 generated task satisfied the condition. However, the verb ``install'', which appeared 3 times in the ChatGPT generated task, did not meet the similarity threshold. In other words, the ``install'' is not a primitive action for robots, thus it needs further decomposition. In addition, the last two steps in the ChatGPT generated task are irrelevant with the assembly process and not considered in the tree expansion.

\begin{table}[htp]
\centering
\caption{Sub-task decomposition using the GPT-3 text-davinci-003 and ChatGPT models. The non-primitive sub-task is ``Install CPU in desktop''. The target task is generated inside the [\enspace]$_Y$ slot. }
\begin{tabular}{l|l}
     \hline
    GPT-3 text-davinci-003 & ChatGPT\\
     \hline
     Source Task &  Source Task\\
     ... & ... \\
(Identical text as Table~\ref{tab:TestPrompt}) & (Identical text as Table~\ref{tab:TestPrompt}) \\
... & ...\\
\\
Target Task: Install CPU in desktop &Target Task: Install CPU in desktop \\in 1 phase, only use the following verb:&in 1 phase, only use the following verb:\\ pick, drop, push, pull, rotate, move, place&pick, drop, push, pull, rotate, move, place\\
Procedures: & Procedures: \\
{[}Step 1. Pick the CPU; & {[}Step 1. Pick up the CPU; \\
Step 2. Place the CPU in the socket;& Step 2. Open the desktop's case \\
 Step 3. Push down the CPU to secure it; &by pulling the tabs or pushing the buttons; \\
Step 4. Rotate the locking arm to secure the CPU;&Step 3. Locate the CPU socket \\
 Step 5. Drop the CPU fan on top of the CPU;&and align the CPU with the socket; \\
 Step 6. Push down the fan to secure it;&Step 4. Carefully place the CPU into the socket;\\
 Step 7. Move the power cable to&Step 5. Secure the CPU in place  \\
 the fan and plug it in.&by fastening any clips or levers;\\
 &Step 6. Close the desktop's case \\
 &by pushing or pulling the tabs or buttons; \\
 &Step 7. Rotate or move any cables \\
 &or connections as necessary.\\
\hline
    \end{tabular}
    \label{tab:SubTestPrompt}
\end{table}

To study the tree expansion, we used the ``Install CPU'' sub-task from the ChatGPT generated task as an example. The source target in the prompt was kept unchanged, and the target task description was specified as ``Install CPU in desktop in 1 phase, only use the following verb: pick, drop, push, pull, rotate, move, place''. For comparison, we also applied this expansion in the GPT-3 text-davinci-003 task even though the corresponding task had already met the condition. The generated sub-tasks are shown in Table~\ref{tab:SubTestPrompt}. After the sub-task decomposition, the verbs in both generated tasks satisfied the condition for primitive tasks. It indicated that there was no need to further expand this branch of the behavior tree. Apart from the condition satisfaction, we can still observe some differences between the generated sub-tasks. The task generated by GPT-3 text-davinci-003 was strictly regulated by our pre-defined verb list while the task from ChatGPT used other verbs outside of our verb list. We think that the ChatGPT might not be capable of understanding the verb restriction in the prompt. On the other hand, the task generated by GPT-3 text-davinci-003 was more succinct and more interpretable by the robots.

\begin{figure}[h]
    \centering
    \includegraphics[width=1\linewidth]{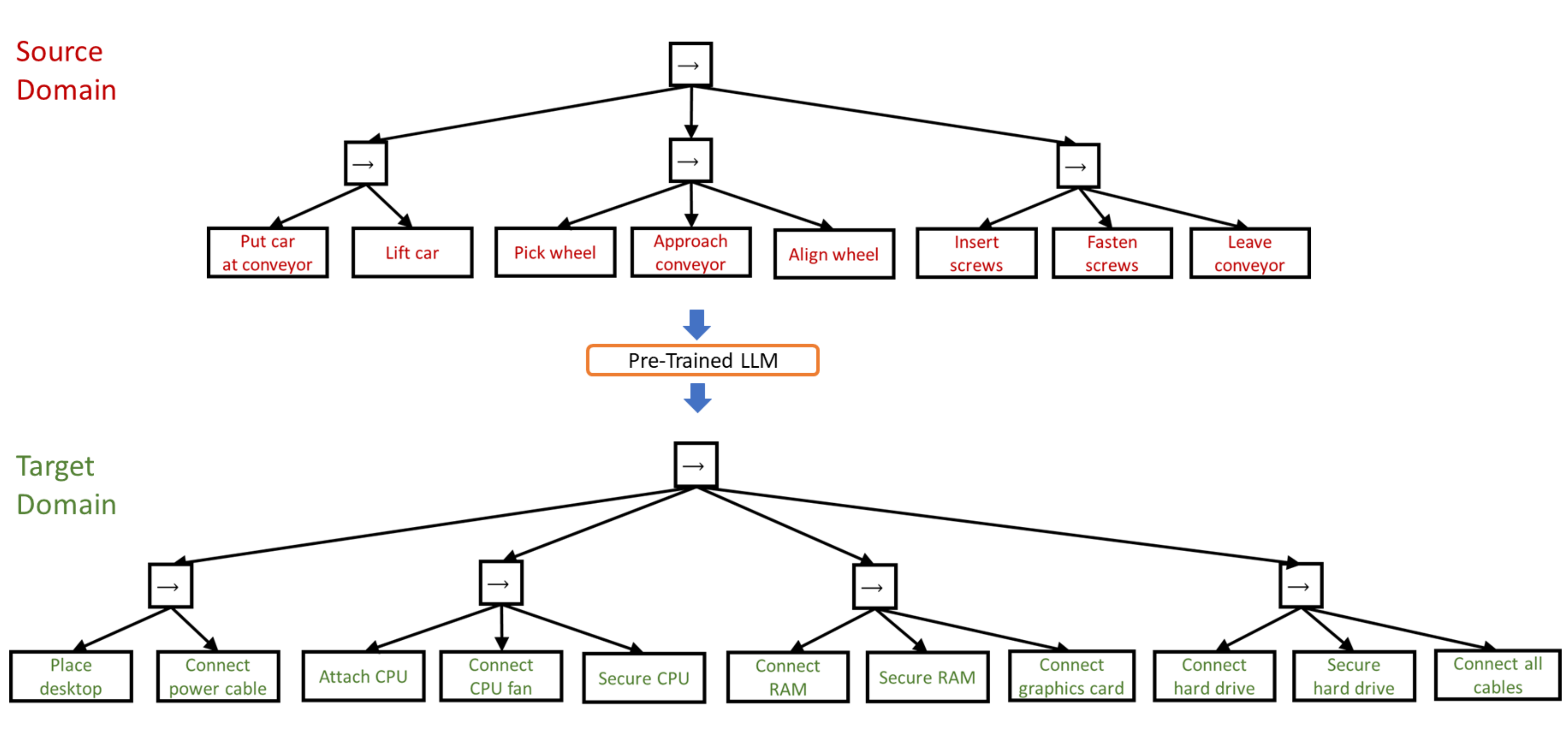}
    \caption{Behavior-tree-based task generation using the GPT-3 text-davinci-003 model. We only keep the core information of primitive tasks in the visualization and refer their details in Table~\ref{tab:TestPrompt}. The source behavior-tree task is vehicle-wheel assembly and the target behavior-tree task is desktop assembly.}   
    \label{fig:finalcase}
\end{figure}

For visualization, a behavior-tree task produced by the GPT-3 text-davinci-003 model is illustrated in Figure~\ref{fig:finalcase}.

\subsection{Ablation Study} 
Next, we focused on the Phase-Step-prompt-based task generation stage and conducted an ablation study. We compared the performance of the non-Phase-Step prompt versus the Phase-Step prompt. The non-Phase-Step is specified as a simple description -- ``Generate a $[\enspace...\enspace]$ task in behavior tree''. Two Phase-step prompts for comparison are set in Table~\ref{tab:MeasurePrompt}. For simplicity, we refer the non-Phase-Step prompt, the wheel-assembly Phase-Step prompt, and the desktop-assembly Phase-Step prompt as \textbf{PS-none prompt}, \textbf{PS-wheel prompt}, and \textbf{PS-desktop prompt}, respectively. The PS-wheel prompt and the PS-desktop prompt included the same number of total steps but differed in terms of tree structure. The PS-wheel prompt was organized in a $(2\rightarrow3\rightarrow3)$ structure whereas the PS-desktop prompt was  shaped in $(3\rightarrow3\rightarrow2)$. 
\begin{table}[htp]
\centering
\caption{We used 3 prompts for comparison -- PS-none prompt, PS-wheel prompt, and PS-desktop prompt. The PS-none prompt is specified as  ``Generate a $[\enspace...\enspace]$ task in behavior tree''. The PS-wheel prompt, and PS-desktop prompt are listed in the table. The $[\enspace...\enspace]$ slot is the place to fill the target task description.}
\begin{tabular}{l|l}
     \hline
    \textbf{PS-wheel prompt} & \textbf{PS-desktop prompt}\\
     \hline
     Source Task &  Source Task\\
Procedures: & Procedures: \\
Phase 1. & Phase 1. \\
Step 1. Put car at a conveyor; &Step 1. Attach the CPU to the motherboard;\\
Step 2. Lift the car.&Step 2. Connect the CPU fan to the motherboard;\\
Phase 2. & Step 3. Secure the CPU with screws.\\ 
Step 1. Pick the wheel; & Phase 2. \\
Step 2. Approach conveyor;& Step 1. Connect the RAM to the motherboard; \\
 Step 3. Align wheel with wheel hub. & Step 2. Secure the RAM with screws; \\
Phase 3. & Step 3. Connect the graphics card to the motherboard. \\
Step 1. Insert screws; & Phase 3. \\
Step 2. Fasten screws; &Step 1. Connect the hard drive to the motherboard; \\
Step 3. Leave the conveyor. & Step 2. Secure the hard drive with screws.
\\
\\
Target Task: {[}\enspace...\enspace{]} & Target Task: {[}\enspace...\enspace{]}\\
Procedures: & Procedures: \\
{[}\enspace\enspace\enspace\enspace\enspace{]}$_Y$ & {[}\enspace\enspace\enspace\enspace\enspace{]}$_Y$ \\
\hline
    \end{tabular}
    \label{tab:MeasurePrompt}
\end{table}

We are primarily interested in the structure of the generated trees since it determines the task modularity and composability. We prefer the generated tasks structured in a well-balanced tree over a sequence. To measure such property, we propose a structure ratio $R$ for behavior trees. We omit condition-check related nodes and count the number of child {\em Action} nodes under each internal node on the second layer. We refer the maximum number of child {\em Action} nodes as $N_{max}$ and the minimum as $N_{min}$. If a behavior tree is only two-layer, in other words, the second layer is just a sequence without any child, then $N_{min}$ is set to be $0$. The structure ratio $R$ is calculated as $R=N_{min}/N_{max}$. In addition, we refer the total number of {\em Action} nodes in the entire behavior tree as $N_{total}$. Take our PS-wheel prompt and PS-desktop prompt as an example, the $N_{min}$ and $N_{max}$ for both are $2$ and $3$, respectively. Thus their measures are both $R=2/3$ and $N_{total}=8$.  

We tested these 3 prompts using GPT-3 text-davinci-003 and ChatGPT. In each prompt test, we used 30 different target task descriptions to fill the ${[}\enspace...\enspace{]}$ space, including 15 in manufacturing assembly, 5 in logistics packaging, 5 in kitchen cooking, and 5 in household cleaning. The evaluation results are shown in Table~\ref{tab:MeasureResult}. All prompts produced multi-step tasks. The total steps $N_{total}$ of the Phase-Step-prompt-generated tasks were close to the $N_{total}$ in the source tasks. Without the Phase-Step prompt, the average of structure ratio $R$ was close to zero. It indicated that the generated task was nearly in a sequence structure rather than a tree structure. After adding the Phase-Step prompt, all $R's$ were close to or above $2/3$, which was the original structure ratio of the source tasks. This measure implied that the target tasks generated by Phase-Step prompts were capable of preserving the tree structure. 

\begin{table}[htp]
\centering
\caption{Quantitative evaluation of generated target task structures. Avg. is the abbreviation for Average.}
\begin{tabular}{lcccc}
\toprule
 &  \multicolumn{2}{c}{GPT-3 text-davinci-003} & \multicolumn{2}{c}{ChatGPT}\\
\midrule
{}   & Avg. $R$   & Avg. $N_{total}$    & Avg. $R$   & Avg. $N_{total}$\\
PS-none prompt  &  0.12 & 5.67   & 0.22  & 6.90\\
PS-wheel prompt &  0.65 & 7.80   & 0.60  & 9.77\\
PS-desktop prompt  & 0.93  &  8.80   & 0.66  & 9.13\\
\bottomrule
\end{tabular}
 \label{tab:MeasureResult}
\end{table}

Apart from lacking modularity, the PS-none prompt also has an issue in interpretation towards robots. Sometimes the PS-none prompt failed to produce a robot-oriented task. For example, when we applied {PS-none prompt} and set to ``Generate a desk assembly task in behavior tree'', GPT-3 text-davinci-003 responded as ``...3. Read Instructions 4. Assemble Desk...'' and ChatGPT responded as ``...Follow the instructions to assemble the bookcase...''. Such ambiguous responses were difficult for robots to understand and did not offer detailed execution steps for robots.

\subsection{Phase-Step Prompt Assessment}
 The large language models can produce robot tasks in a variety of domains. However, there lacks a universal metric to evaluate generated robot tasks spanning among different domains.  To provide a useful metric to evaluate robot-task operations, we narrow down our assessment scope to the assembly domain only~\cite{nist817581}. One key type of assembly operation is the part-mating operation that directly affects the task executability. In this assessment, we count the total number of part-mating operations in a generated task and denote this number as $N_{mate}$. The source task in the PS-wheel prompt has only 1 part-mating operation, whereas the source task in the PS-desktop prompt indicates 5 part-mating operations.

Since we have already obtained 15 assembly tasks for each prompt in Section 4.3, we considered these tasks and performed our proposed evaluation metric on them. The part-mating operation evaluation results are presented in Table~\ref{tab:AssemblyResult}. Tasks generated by the PS-wheel prompt tended to have fewer part-mating operations than tasks generated by the PS-desktop prompt. 

\begin{table}[htp]
\centering
\caption{Part-mating operation counting in the assembly tasks. Avg. is the abbreviation for Average.}
\begin{tabular}{lcccc}
\toprule
 &  \multicolumn{2}{c}{GPT-3 text-davinci-003} & \multicolumn{2}{c}{ChatGPT}\\
\midrule
{}   & Avg. $N_{mate}$   & Avg. $N_{total}$    & Avg. $N_{mate}$   & Avg. $N_{total}$\\
PS-wheel prompt &  1.87 & 7.80   & 3.20  & 10.80\\
PS-desktop prompt  & 5.33 & 8.87   & 5.73  & 10.00\\
\bottomrule
\end{tabular}
 \label{tab:AssemblyResult}
\end{table}

For example, we set one target task description to be ``Projector manufacturing assembly'' and tested using GPT-3 text-davinci-003. The task produced by the PS-wheel prompt resulted only in one part-mating operation, that was lens-projector body mating. Meanwhile, the task created by the PS-desktop prompt contained 5 part-mating operations -- lens-projector body mating, fan-projector body mating, lamp-projector body mating, circuit board-projector body mating, and power cord-projector body mating. 

In summary, we discovered that the level of details in prompt affects the quality of generated tasks. we favor prompts with more informative description so that we can obtain more detailed procedures in the target tasks. 

\subsection{Limitation}
During our tests, we noticed that the large language models have limited capability when dealing with {\em rare} tasks. For example, we used the PS-desktop prompt and set the target task description to be ``Vince Lombardi Trophy crafting'', which is the championship trophy in Super Bowl. The GPT-3 text-davinci-003 successfully generated a task, especially mentioning the football shaped design in the task. However, the ChatGPT frequently failed to answer it. Its responses included ``there is no clear connection between the source task and the target task'', ``process and materials are not publicly available'', ``beyond my knowledge cutoff date of 2021'', or ``Sorry, as a language model, I am not able to perform physical tasks''. Another example is the Atlas robot assembly task. The ChatGPT successfully created an assembly task, notably including its featuring hydraulic actuators. But the GPT-3 text-davinci-003 appeared to be ignorant of the speciality of Atlas robots and only stated using servo motors in the assembly.

\section{Conclusions and Discussions}
This paper presented a robot behavior-tree-based task generation utilizing the state-of-the-art large language models. 
It brought together the new-task-representation trend in robotics and the generative AI technique in natural language processing. By enabling cross-domain task transfer, this approach reduces end-users' work load in designing new robot tasks.

In order to apply this approach to real-world robots, further studies are still required. For example, although the generated robot-manufacturing tasks look reasonable and executable, it does not take the complex factory environment into consideration. In industry applications, the design of robot tasks has to follow strict safety regulations. How to incorporate such safety regulations into the task generation remains an open question.
\bibliography{sample-ceur}

\end{document}